\documentclass[acmsmall,nonacm]{acmart}
\AtBeginDocument{%
  }

\setcopyright{acmlicensed}
\copyrightyear{2024}
\acmYear{2024}
\acmDOI{XXXXXXX.XXXXXXX}
\acmConference[WWW]{The Web Conference 2025}{April 28 - May 2, 2025}{Sydney, Australia}
\acmISBN{978-1-4503-XXXX-X/18/06}
\usepackage{float}
\usepackage{hyperref}
\usepackage{enumitem}
\usepackage{graphicx}
\usepackage[most]{tcolorbox}
\usepackage[skip=3pt]{caption}
\usepackage{longtable}
\usepackage{adjustbox}
\usepackage{booktabs}
\usepackage{placeins}

\begin{document}

\title[A Platform for Efficient Concern Classification]{A Platform for Investigating Public Health Content \\ with Efficient Concern Classification}

\author{Christopher Li}
\email{christopher.li@duke.edu}
\orcid{}
\affiliation{%
  \institution{Duke University}
  \country{USA}
}

\author{Rickard Stureborg}
\email{rickard.stureborg@duke.edu}
\orcid{}
\affiliation{%
  \institution{Duke University}
  \country{USA}
}

\author{Bhuwan Dhingra}
\email{bdhingra@cs.duke.edu}
\orcid{}
\affiliation{%
  \institution{Duke University}
  \country{USA}
}

\author{Jun Yang}
\email{junyang@cs.duke.edu}
\orcid{}
\affiliation{%
  \institution{Duke University}
  \country{USA}
}

\renewcommand{\shortauthors}{Li et al.}

\begin{abstract}
  A recent rise in online content expressing concerns with public health initiatives has contributed to already stalled uptake of preemptive measures globally.
  Future public health efforts must attempt to understand such content, what concerns it may raise among readers, and how to effectively respond to it.
  To this end, we present ConcernScope, a platform that uses a teacher-student framework for knowledge transfer between large language models and light-weight classifiers to quickly and effectively identify the health concerns raised in a text corpus.
  The platform allows uploading massive files directly, automatically scraping specific URLs, and direct text editing.
  ConcernScope is built on top of a taxonomy of public health concerns.
  Intended for public health officials, we demonstrate several applications of this platform:
    guided data exploration to find useful examples of common concerns found in online community datasets,
    identification of trends in concerns through an example time series analysis of 186{,}000 samples, and 
    finding trends in topic frequency before and after significant events.
\end{abstract}




\maketitle
\section{Introduction}
Understanding public concerns about health interventions is a critical challenge for public health practitioners, particularly in the domain of vaccine hesitancy. 
In recent years, vaccine concerns have posed growing challenges to public health as vaccine rates have stalled~\citep{neely2022vaccine,gur2019vaccine}. 
Identifying trends in such concerns would help inform public health campaigns to address specific topics and guide future efforts~\citep{stureborg2024phd}.
Prior research in this space largely focuses on detecting sentiment---positive, neutral, or negative---toward vaccines on social media~\citep{twitter2021covidsentiment, vaccine2017sentiment, youtube2019vaccinesentiment}.
However, many of these studies do not analyze the specific themes underlying these sentiments and few track how these concerns change over time. 
There are also formal reporting systems and social listening tools within the public health field, such as the Vaccine Adverse Event Reporting System (VAERS) and the World Health Organization's (WHO) Early AI-supported Response with Social listening (EARS) platform~\citep{vaers, ears}. 
Unfortunately, many such platforms (including EARS) have been shut down or are no longer maintained, partly due to increased restrictions and pricing of social media data access~\citep{twitterprices} and shifting prioritization of government policies and funding~\citep{nih2025cuts, kentucky2025dogecuts}.

Motivated by the need to monitor concerns towards vaccines, we propose ConcernScope, a platform that can quickly and effectively identify vaccine concerns in a corpus. 
This application is intended to quickly and accurately classify large amounts of text data, summarize the results, and return vetted information relevant to the concerns. 
We further demonstrate additional applications, including identifying trends in vaccine concerns through time series analysis and examining shifts in concerns surrounding significant events. 

We build directly off of prior work on classifying the landscape of public health concerns surrounding vaccination, utilizing the \emph{VaxConcerns} taxonomy~\citep{vaxconcerns}, a disease-agnostic taxonomy of concerns that may drive people towards hesitancy.
This taxonomy organizes concerns into two levels, one of broad granularity with concern categories such as \textit{Health Risks} and another of finer granularity with specific claim categories such as vaccines having \textit{Harmful Ingredients} or \textit{Specific Side-Effects}. 
It is composed of 5 parent categories and 19 child classes each associated with only one parent category. 
To efficiently classify text using VaxConcerns, we fine-tune a BERT model by distilling labeled data from GPT-4, leveraging GPT-4's high-quality annotations as a foundation for the task ~\citep{distillation2024}.
We focus primarily on GPT-4 as prior work found it achieved strong performance among several closed and open-source LLMs~\citep{zhu2024hierarchicalmultilabelclassificationonline}.

Classification on the hierarchical taxonomy's label-set can include multiple concerns in a single passage of text, and we therefore build on previous work on hierarchical multilabel classification task~\cite{hierarchical2023multilabel}.
Our models produce independent binary prediction (``present'' or ``not present'' in the text) for every node in the VaxConcerns taxonomy.
These predictions must be made for both the parent and child categories, since a text can invoke the broad category (e.g. \textit{Lack of Benefits}) with and without a specific child rationale (e.g. \textit{Existing Alternatives}). 
For example, consider the YouTube comment: ``\textit{I don't need the vaccine! No reason to get it}'', which clearly invokes the parent label \textit{Lack of Benefits} without invoking any of the child labels.

In the remainder of the paper, we first discuss, in Section~\ref{sec:framework}, our method for building such a hierarchical multilabel classifier for ConcernScope.
Then, Section~\ref{sec:demo} describes our web-based ConcernScope platform to be demonstrated.
Finally, Section~\ref{sec:pilot} presents a case study of 186,000 samples from 9 blogs that regularly attack vaccination as a practice, demonstrating how ConcernScope can capture trends in concerns surrounding vaccination over time.

\section{Classifying Vaccine Concerns}\label{sec:framework}

Classifying text to identify vaccine concerns is a difficult task even for human annotators~\cite{stureborg2023interface},
but LLMs have been shown to perform well under zero-shot settings when labels are well defined~\cite{stureborg2024phd}.
However, using LLMs directly for this task remains prohibitive due to computational costs and reasonable latency requirements~\cite{llama2023benchmarking}.
Therefore, we explore an alternative approach that first uses LLMs to label a training dataset;
then, using this training dataset, we finetune a less expensive model ~\citep{hinton2015} (such as BERT) to be used by ConcernScope at runtime.
Ideally, this knowledge transfer process with synthetic data generation retains the LLMs' high performance while requiring much less computation.

Although ConcernScope needs only one hierarchical multilabel classifier in the end,
during the process of training this classifier, we employ a second vaccine relevance classifier.
The purpose of this classifier is to quickly filter out text that is not vaccine-related from more expensive labeling of vaccine concerns.
Both classifiers are trained using a similar approach, leveraging knowledge transfer from LLMs.

The approach of using LLMs to create a training dataset for finetuning less expensive models requires highly accurate LLMs.
In their technical report, \citet{gpt-4} show that GPT-4 demonstrates strong reading comprehension, as it exhibits human-level performance on many academic exams and advances the SOTA on several NLP benchmarks.
Given our tasks' complexity and the need for high accuracy, we therefore use GPT-4 for generating and labeling the training dataset in a zero-shot setting.


\subsection{Data Preparation}
Our primary dataset is composed of $186{,}000$ passages from various websites (Table~\ref{tab:dataset}). These websites have both articles attacking vaccination as well as articles not related to vaccines. 
From this dataset, we sample $500$ passages and label each of them for vaccine relevance by hand.
We also sample $200$ passages and annotate them according to the VaxConcerns taxonomy by hand.
We consider these the gold labeled datasets for the two respective tasks.

\begin{table}[]
    \centering
    \begin{tabular}{l r}
        \toprule
        \textbf{Website name} & \textbf{\# Articles} \\
        \midrule
        Vaxxter & 989 \\
        Green Med Info & 2819 \\
        Modern Alternative Mama & 912 \\
        Thinking Moms Revolution & 1031 \\
        Age Of Autism & 4698 \\
        Australian Vaccination-risks Network & 300 \\
        The Vaccine Reaction & 312 \\
        Focus For Health & 185 \\
        Vactruth & 1117 \\
        \bottomrule
    \end{tabular}
    \caption{Our analysis spans a sample collection of articles sourced from 9 blogs. Approximately 12,000 articles are included in the analysis, which were split into 186,000 individual passages.}
    \Description{A summary table presenting the number of articles from 9 different websites. Several of these websites have been actively publishing articles for more than a decade.}
    \label{tab:dataset}
\end{table}

\subsection{Relevance Classifier}
\subsubsection{Evaluating GPT-4.} First, we evaluate GPT-4's ability to determine whether a given text is related to vaccines or not. With minimal prompt engineering to avoid overfitting, we prompt GPT-4 on each of the $500$ passages manually labeled for vaccine relevance. 
We use the following prompt:
\begin{tcolorbox}[colback=gray!5!white, colframe=gray!75!black, boxsep=1mm, left=1mm, right=1mm, top=1mm, bottom=1mm]
\textit{You will be given a small paragraph of text. Please return whether the text is relevant to vaccines. Text is vaccine-related if it mentions vaccines in some way, indirectly or directly. Note that the sentences may be vaccine relevant even if there aren't any keywords like "vaccine" or "vaccination". Think carefully about your answer as this task is important, then return a `Yes' or `No' indicating if the paragraph discusses vaccination. \textbackslash n Paragraph input: \{paragraph\}}
\end{tcolorbox}

GPT-4 performs extremely well at this binary classification task (Table~\ref{tab:relevance_models}). 
Nearly all paragraphs containing a vaccine-related keyword are labeled as vaccine-related by both GPT and human annotators. 
This task is fairly straightforward, as the presence of a vaccine keyword in the text strongly correlates with a positive classification.

\begin{table}[H]
    \centering
    \begin{tabular}{l c r c l}
        \toprule
        \textbf{Model} & \textbf{Accuracy} & \textbf{Precision} & \textbf{Recall} & \textbf{F1} \\
        \midrule
        \textit{GPT-4} & \textbf{0.968} & \underline{0.981} & \textbf{0.969} & \textbf{0.975} \\
        \midrule
        Keyword & 0.929 & \textbf{0.993} & 0.895 & 0.941 \\
        \midrule
        XGBoost & 0.911 & 0.904 & \underline{0.963} & 0.933 \\
        \midrule
        BERT & \underline{0.954} & 0.969 & 0.960 & \underline{0.964} \\
        \bottomrule
    \end{tabular}
    \caption{Model performance on the vaccine relevance classification task.
    Best performances are bolded and second bests are underlined.
    All models (except GPT-4) are trained using synthetic data labeled by GPT-4, which achieves the highest performance on the test set. 
    The BERT classifier ranks second, offering a highly accurate alternative to GPT-4 that requires significantly less compute.}
    \Description{This table presents the performance of the keyword, XGBoost, and BERT models on the relevance task.}
    \label{tab:relevance_models}
\end{table}

\subsubsection{Annotating Relevance using GPT-4}
Due to the cost of using OpenAI's API, we randomly sample $10{,}000$ passages from the full dataset of $186{,}000$ passages to annotate with GPT-4 for vaccine relevance.

\subsubsection{Training a Relevance Classifier}
Initially, we experiment with several baseline models. 
The naive keyword-based classifier performs modestly but struggles to correctly classify vaccine-related passages that contained uncommon phrases or rare words indicative of a positive classification. 
Using grid search cross-validation, XGBoost shows slight improvement but still falls short when evaluated on the gold labeled dataset.

Finetuning a BERT model on GPT-4-annotated data achieves performance comparable to GPT-4. 
We append a dropout layer, a linear layer, and an output layer to the pretrained model, and then optimize it using backpropagation with cross-entropy loss. 
This approach demonstrates effective knowledge transfer for the relevance task, as shown by BERT’s performance on the gold-labeled dataset (Table~\ref{tab:relevance_models}).

Testing during development included synthetically generated vaccine-related texts which intentionally exclude common vaccine-related keywords.
Anecdotally, BERT's performance on such texts was strong, and we include further details on generation in Appendix~\ref{appendix:testing_relevance}.

\subsection{Multilabel Classifier}
\subsubsection{Evaluating GPT-4.} We experiment with two different prompting approaches for this classifications task: an all-in-one and an individual prompting approach. In both approaches, we provide the full taxonomy and its definitions. 
In the all-in-one approach, we ask for all $24$ classifications at once. 
For the individual prompting approach, we only ask for the classification for one of the classes, sampling $100$ times and then choosing a threshold value to maximize F1. 
Intuition may suggest that isolating each class gives GPT-4 a better chance to decide. 
However, the individual prompting approach is significantly worse, as shown in Table~\ref{tab:multilabel_prompting}.

\begin{table}[H]
    \centering
    \begin{tabular}{l c r c}
        \toprule
        \textbf{Prompting Strategy} & \textbf{Precision} & \textbf{Recall} & \textbf{F1} \\
        \midrule
        All-in-one & 0.75 & 0.68 & 0.69 \\       
        \midrule
        Individual & 0.46 & 0.68 & 0.51 \\
        \bottomrule
    \end{tabular}
    \caption{Comparison of all-in-one and individual prompting strategies for GPT-4 for multilabel classification, using sample-averaged metrics. The all-in-one prompting strategy performs better than the individual prompting strategy on all metrics. This is despite the individual prompting strategy having context about the full taxonomy and all possible labels, thereby supporting previous findings in \citet{zhu2024hierarchicalmultilabelclassificationonline}.}
    \Description{This table compares the performance of two prompting strategies, ``All-in-one'' and ``Individual,'' across precision, recall, and F1-score metrics. The values are averaged over samples.}
    \label{tab:multilabel_prompting}
\end{table}

\subsubsection{Annotating VaxConcerns Labels with GPT-4.}
Starting with all $186{,}000$ passages, we first apply the BERT relevance classifier to select only those are vaccine relevant.
Among these, we randomly sample $10{,}000$ passages and prompt GPT-4 to produce the VaxConcerns labels.

\subsubsection{Training BERT for multilabel classification.}
The BERT multilabel classifier is similar to the relevance classifier, with a dropout, linear, and output layer added to the end of a pre-trained BERT model. 
We train it on the $10{,}000$ GPT-4 labeled vaccine relevant passages, using binary cross-entropy loss with backpropagation.

Due to significant class imbalance, the baseline model tends to avoid classifying some classes as positive. 
To address this, we weight the loss function, assigning higher weights to minority classes and lower weights to majority classes ~\citep{weighted2020loss}. 
For example, if a class has 100 positive and 300 negative examples, the positive weight would be $\frac{300}{100} = 3$, effectively treating the dataset as if it contains 300 positive examples. 
We experiment with different weighting schemes, including clipping and logarithmic scaling (Figure~\ref{tab:multilabel_weighting}).
As a result, recall improves because the model correctly identifies more instances of minority classes, but this comes at the cost of precision, as the classifier also generates more false positives.
Among the weighting methods tested, logarithmic scaling achieves the best balance between maintaining accuracy on majority classes while improving prediction for minority classes, so we adopt this model for ConcernScope.

\begin{table}[H]
    \centering
    \begin{tabular}{l c r c}
        \toprule
        \textbf{Weighting Scheme} & \textbf{Precision} & \textbf{Recall} & \textbf{F1} \\
        \midrule
        
        baseline & \textbf{0.74} & 0.49 & 0.56 \\       
        \midrule
        clamp at 3 & 0.73 & 0.52 & 0.58 \\
        \midrule
        clamp at 10 & 0.65 & 0.50 & 0.54 \\
        \midrule
        clamp at 30 & 0.69 & 0.55 & 0.59 \\
        \midrule
        clamp at 100 & 0.67 & 0.51 & 0.55 \\
        \midrule
        no clamp & 0.69 & 0.51 & 0.56 \\
        \midrule
        log1p & 0.70 & \textbf{0.56} & \textbf{0.60} \\
        \bottomrule
    \end{tabular}
    \caption{Weighting scheme selection leads to a \(\sim\)7\% improvement in recall. This is due to the impact of class imbalance when training the multilabel classifier. Each scheme assigns weights to classes based on their distribution. The baseline does not use any loss weighting, while alternative schemes either clamp the weights, leave them unchanged, or apply a logarithmic adjustment. log1p achieves the best performance, effectively balancing predictions for minority classes while maintaining strong accuracy for majority classes.}
    
    \Description{This table evaluates various weighting schemes on the downstream multilabel classification task, reporting precision, recall, and F1-score for each scheme. The baseline and adjusted schemes (e.g., clamping at different thresholds or applying logarithmic transformation) are compared.}
    \label{tab:multilabel_weighting}
\end{table}

We further evaluate this multilabel classifier on the $500$-passage relevance task; results are summarized in Table~\ref{tab:multilabel_relevance}.
Using the multilabel classifier, we would consider a passage vaccine relevant if any of its classes are labeled as positive. 
While the multilabel classifier can be used in this fashion, their performance is hindered by cases where a passage is vaccine-related but does not explicitly express concerns.
It is possible that the classifier imagines some concerns, since the precision also suffers.

\begin{table}[H]
    \centering
    \begin{tabular}{l c r c l}
        \toprule
        \textbf{Classifier} & \textbf{Accuracy} & \textbf{Precision} & \textbf{Recall} & \textbf{F1} \\
        \midrule
        Relevance & \textbf{0.954} & \textbf{0.969} & \textbf{0.960} & \textbf{0.964} \\
        \midrule
        Multilabel Baseline & 0.643 & \underline{0.764} & 0.641 & 0.697 \\
        \midrule
        Multilabel Weighted & \underline{0.649} & 0.759 & \underline{0.663} & \underline{0.707} \\
        \bottomrule
    \end{tabular}
    \caption{Multilabel classifiers perform worse on predicting vaccine relevance of a given input text as compared to a dedicated relevance classifier. This is mainly due to high false positive rates caused by the many independent binary predictions made across taxonomy nodes by the multilabel classifiers. Simple weighting improves performance slightly but is still far worse than a dedicated relevance classifier.}
    \Description{This table presents the performance of the relevance, multilabel baseline, and multilabel weighted classifiers on the relevance task.}
    \label{tab:multilabel_relevance}
\end{table}

\section{Demonstration of ConcernScope} \label{sec:demo}
\begin{figure*}[t]
\begin{minipage}{0.75\textwidth}
    \centering
    \includegraphics[width=1\linewidth, trim=1.5in 4in 1.5in 0mm, clip]{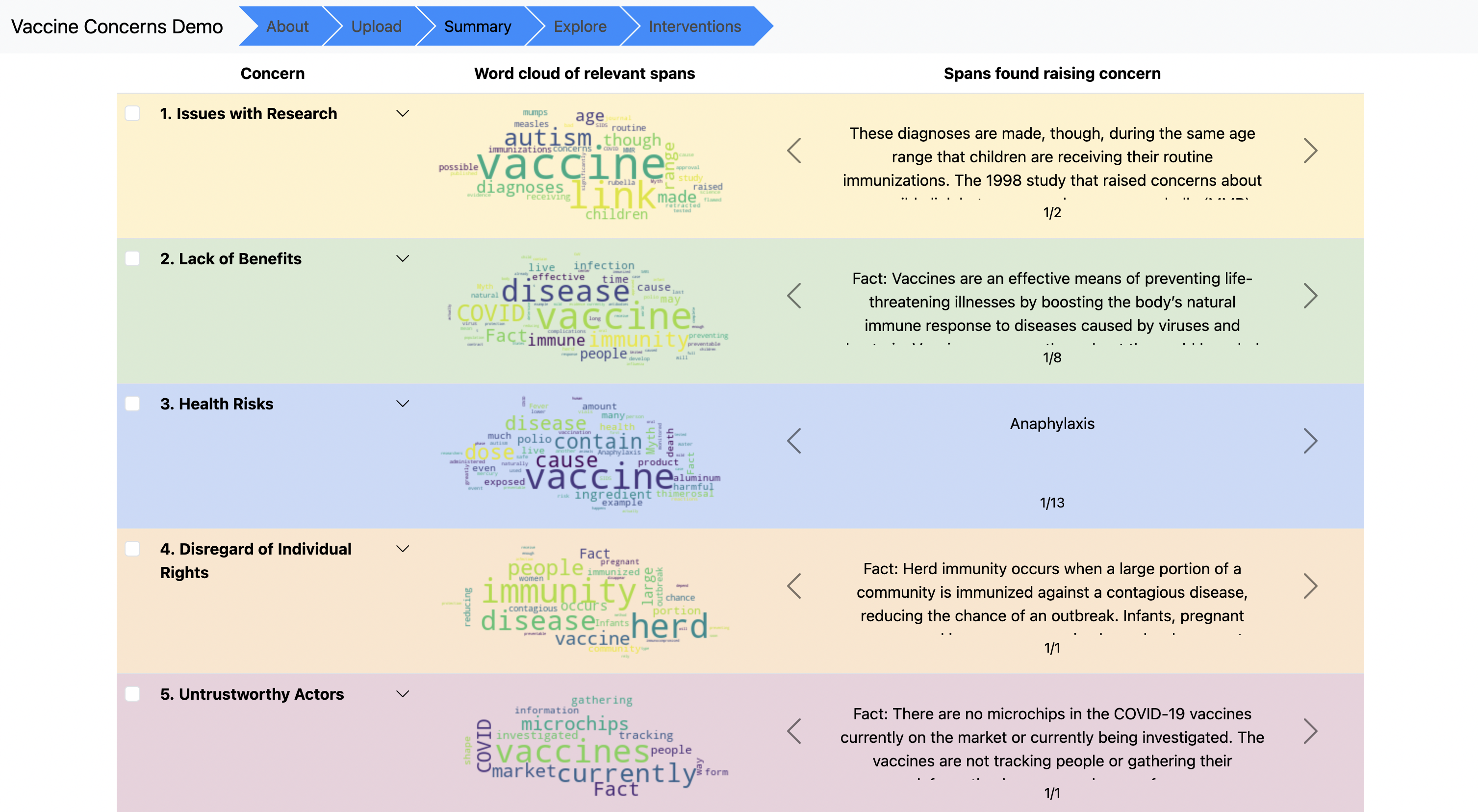}
    \caption{The Summary page of ConcernScope allows guided analysis of common trends and key example spans. 
    Samples are organized by concern category according to the VaxConcerns taxonomy, and word clouds display keyword frequencies within each category. 
    By clicking through the samples, users can explore the raw data and find common trends or entities from their uploaded source data. 
    Under the broad \textit{Health Risks} category, we find that common keywords include \textit{``thimerosal''} and \textit{``aluminum''}, two specific ingredients claimed as harmful by the source blogs.}
    \Description{A screenshot of the summary page, which displays a word cloud and specific examples for each concern.}
    \label{fig:summary}
\end{minipage}
\hfill
\begin{minipage}{\textwidth}
    \centering
    \includegraphics[width=1\linewidth, trim=1.5in 0mm 1.5in 0mm, clip]{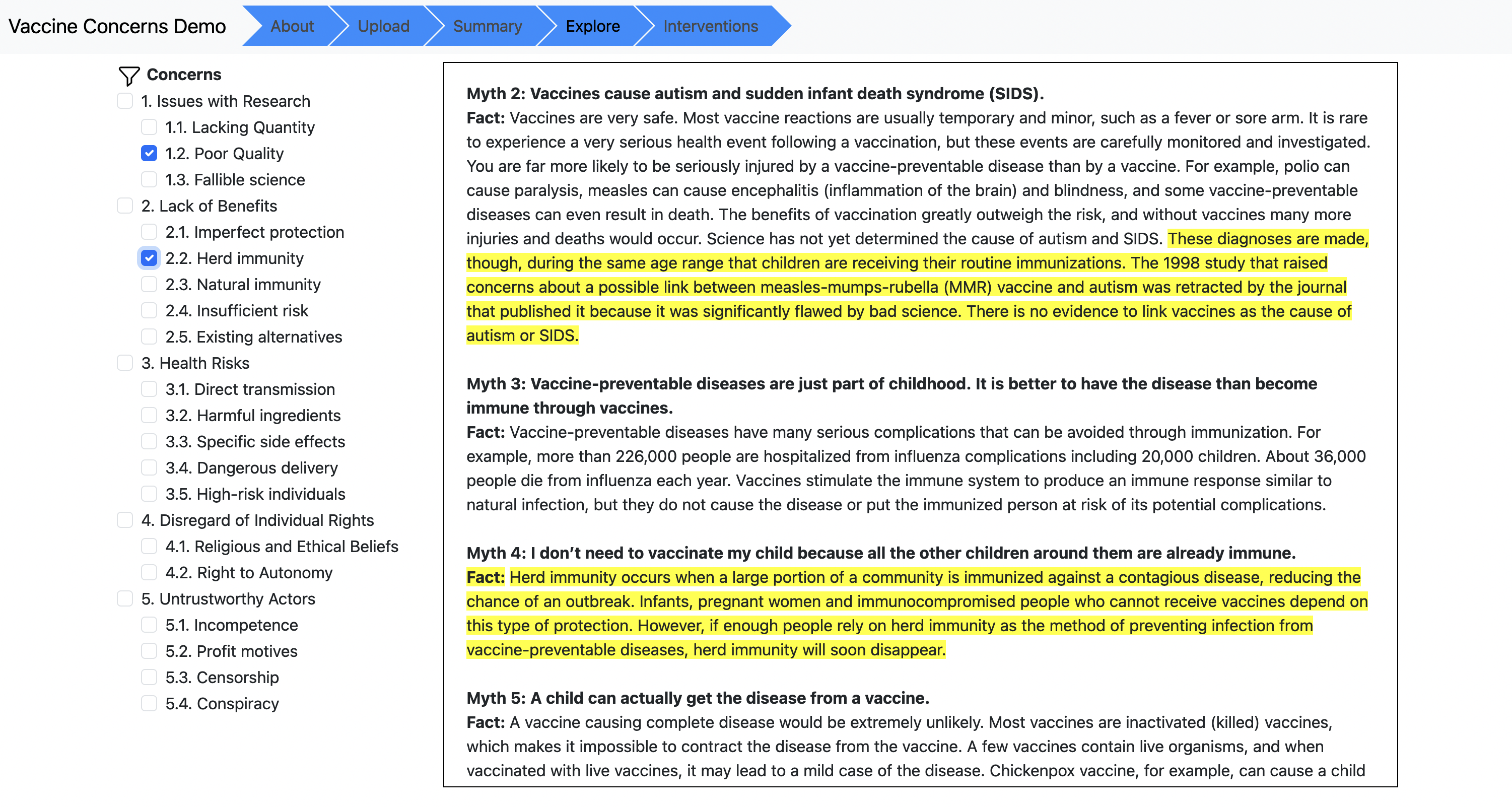}
    \caption{On the Explore page of ConcernScope, users can inspect specific concerns in context. 
    Users can filter the original article text by selecting concerns from the VaxConcerns taxonomy on the left, which highlights corresponding spans within the article. 
    In this example, selecting the \textit{Poor Quality} concern highlights text addressing retracted studies that sparked factually incorrect claims regarding vaccines. 
    This functionality allows users to trace specific themes back to their original language.}
    \Description{A screenshot of the explore page, which displays classified text in the context of the webpage.}
    \label{fig:explore}
\end{minipage}
\end{figure*}

ConcernScope features a web-based interface that allows users such as public health workers to quickly classify and analyze large corpora for vaccine concerns. 
The interface supports various methods for input data: the user can upload a file, input text, or specify a URL.

After uploading data, ConcernScope employs the multilabel classifier to classify the text.
For example, after submitting the URL \href{https://www.aaaai.org/tools-for-the-public/conditions-library/allergies/vaccine-myth-fact}{https://www.aaaai.org/...}, which points to an article describing vaccine myths and facts,
we can navigate to the summary tab to view the classification results, as shown in Figure~\ref{fig:summary}.
The summary page features a word cloud for each concern in the taxonomy, along with the option to browse specific examples. Additionally, the explore tab allows users to observe where the classifier has identified specific concerns in context.
Figure~\ref{fig:explore} illustrates the webpage with concerns such as poor quality and herd immunity.

ConcernScope also provides tailored interventions to address specific vaccine-related concerns. 
These interventions consist of 50 patient and expert handouts sourced from immunize.org. 
The intervention page features a text box where users can input a single concern query. 
Upon submission, the passage is classified, and the website displays the interventions that most closely match the distribution of the concern query, using Jaccard similarity. 
As future work, we plan to add more interventions, hand-classify these to ensure accuracy, and explore alternative matching criteria, such as framing the task as a maximum coverage problem where we seek to address all concerns using a small collection of interventions.

\section{Pilot Study of ConcernScope for Trend Analysis}\label{sec:pilot}
To further demonstrate the utility of ConcernScope, we analyze the full $186{,}000$-sample dataset. 
Each article is divided into multiple passages that are classified individually. 
Then, results are aggregated by article, taking the maximum along each class. 
Essentially, if any passage in an article is classified as relevant to a concern, the entire article is considered relevant to that concern.
Demonstrating the platform's speed and cost-effectiveness, all $186{,}000$ samples were classified in just 15 minutes on a P100 GPU.

\begin{figure}[H]
    \centering
    \includegraphics[width=1\linewidth]{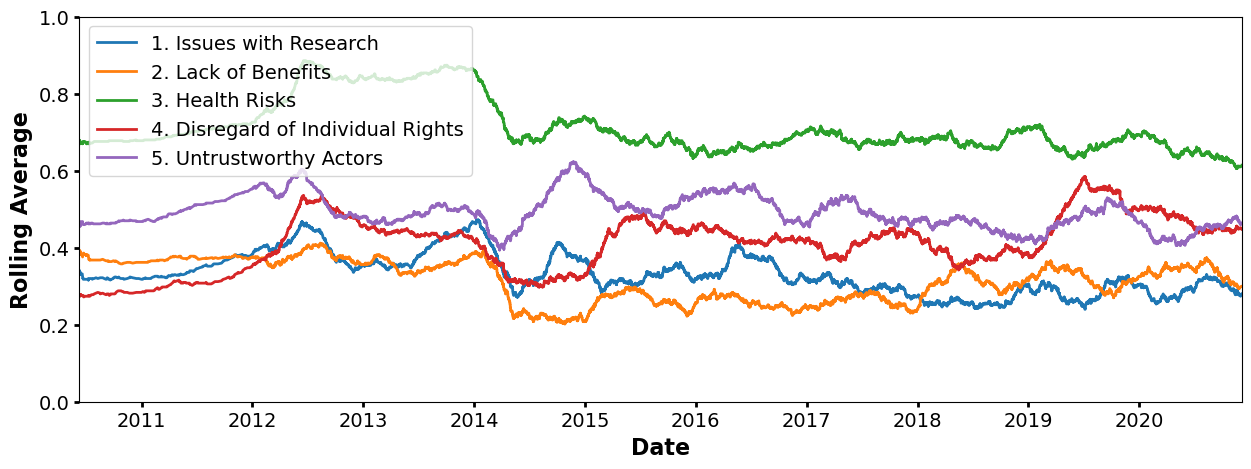}
    \caption{Different vaccine concerns fluctuate over time, revealing shifting public focus. 
    This figure shows the rolling average (500-article window) of the proportion of articles mentioning each of five parent concerns from the VaxConcerns taxonomy. 
    Notably, \textit{Health Risks} dominated early discourse, while \textit{Disregard of Individual Rights} spiked around 2012, 2015, and again near 2019.}
    \Description{This figure displays the rolling average (500-article window) of articles associated with the 5 parent concerns: ``\textit{Issues with Research},'' ``\textit{Lack of Benefits},'' ``\textit{Health Risks},'' ``\textit{Disregard of Individual Rights},'' and ``\textit{Untrustworthy Actors}.''}
    \label{fig:rolling_avg}
\end{figure}

\begin{figure}[H]
    \centering
    \includegraphics[width=1\linewidth]{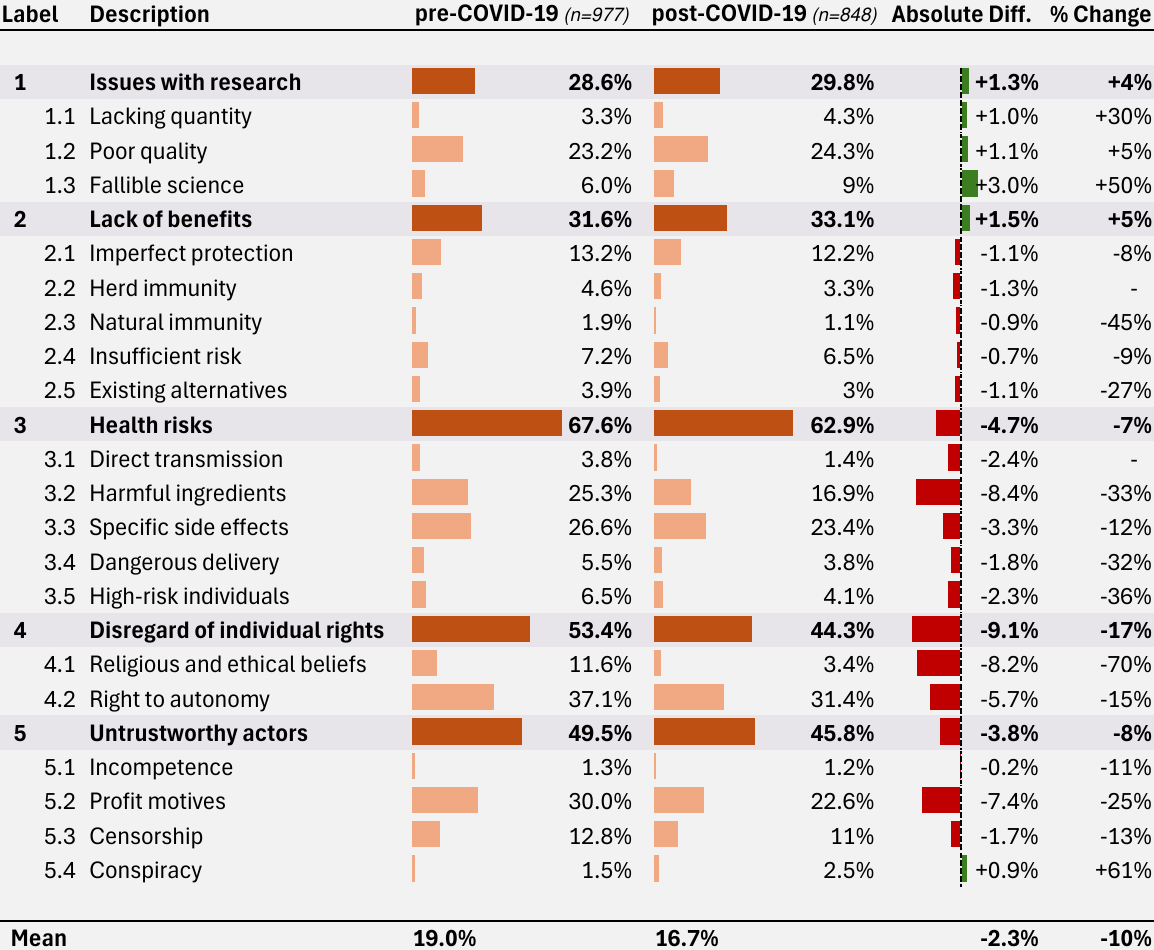}
    \caption{Public vaccine concerns shifted significantly after the onset of COVID-19.
    This figure compares the proportion of articles referencing each concern before and after the announcement of COVID-19 on March 1, 2020.
    \textit{Health Risks and Disregard of Individual Rights} saw substantial declines—particularly in subcategories like \textit{Religious and Ethical Beliefs} (–70\%) and \textit{Harmful Ingredients} (–33\%).
    Conversely, \textit{Conspiracy} concerns rose by 61\%, reflecting growing distrust.
    By analyzing changes in concern prevalence over time, ConcernScope helps users identify emerging trends and adapt public health messaging accordingly.}
    \Description{This figure compares the proportion of vaccine-related concerns in articles published before and after the announcement of COVID-19 on March 1, 2020. It reflects how public concerns surrounding vaccines evolved in response to the pandemic.}
    \label{fig:before_after}
\end{figure}

For each article, we take the average of the classifications over the past $500$ articles and plot this rolling average. 
This analysis can offer valuable guidance for public health researchers and officials. 
For instance, Figure~\ref{fig:rolling_avg} illustrates a general decrease in all parent concerns in 2014, with a notable rise in concerns about the disregard of individual rights in 2019.
We also conduct an analysis of articles published in the months immediately before and after March 1, 2020, to examine how public concerns evolved in response to the emergence of COVID-19. 
The results are shown in Figure~\ref{fig:before_after}. 

While the pilot is still limited in scale and scope, it demonstrates the potential use of the platform. 
ConcernScope effectively identifies vaccine concerns and trends, providing actionable insights for public health officials. 
Its applications, such as better messaging~\citep{stureborg2024tailoring} and analyzing longitudinal sentiment shifts, could play a crucial role in addressing vaccine concerns as a contributor to hesitancy. 
Future work will focus on refining ConcernScope's user experience and functionality to maximize the platform's utility in public health strategies.

\bibliographystyle{ACM-Reference-Format}
\bibliography{base}

\appendix

\section{Further Testing of the BERT Relevance Classifier}
\label{appendix:testing_relevance}
As a sanity check, we briefly assess BERT's performance on adversarial passages—vaccine-related texts that intentionally exclude common vaccine-related keywords. 
This highlights the classifier's robustness in recognizing relevant content even in challenging scenarios, where keyword-based approaches might fail. 
For example, BERT correctly returns a positive classification for this adversarial passage:
\begin{tcolorbox}[colback=gray!5!white, colframe=gray!75!black, boxsep=1mm, left=1mm, right=1mm, top=1mm, bottom=1mm]
\textit{In recent discussions about global health, various strategies have been proposed to address infectious diseases and enhance community well-being. Researchers are actively exploring ways to boost the immune system and develop new technologies for disease prevention, contributing to a healthier and more resilient world.}
\end{tcolorbox}

\section{Prompts for GPT-4 Multilabel Classification} \label{appendix:multilabel_prompt}
The prompt for the all-in-one prompting approach is as follows:
\begin{itemize}[leftmargin=*]\itshape
    \item[]
    You are a healthcare expert helping to determine whether a passage includes vaccine concerns. You have a deep understanding of vaccine-related topics and are capable of providing accurate assessments regarding specific vaccine concerns mentioned in the passage. 

    You will be given a passage and a set of vaccine concerns in a hierarchical order, labeled as "VaxConcerns\_1", "VaxConcerns\_1.1", … , "VaxConcerns\_5.4". You will have the definition for each of the labels.

    Vaccine Concerns:
    \begin{itemize}
        \item VaxConcerns\_1: "Issues with Research" - Criticism of the research of vaccines, whether attacking the quantity or quality of existing research, or generally making the point that science and studies cannot tell us things for certain. Equating inconclusive or bad science to not trusting vaccines.
        \begin{itemize}
            \item VaxConcerns\_1.1: "Lacking Quantity" - Argues that there is not enough research to answer a specific question or concern regarding vaccines. In this view, the implied solution is to conduct more scientific experiments.
            \item VaxConcerns\_1.2: "Poor Quality" - Attacks elements of some existing piece of vaccine research to invalidate it or cast doubts on its results. The implied solution is to redo the experiment or analysis to fix the issues of quality.
            \item VaxConcerns\_1.3: "Fallible science" - Raises doubt in knowledge regarding vaccines based on the fact that you can never be 100\% sure of research conclusions. This view implies that more or better experiments will not solve the issue.
        \end{itemize}
        \item \ldots
    \end{itemize}
    Please read the passage and determine whether the specific concern about the vaccine is mentioned. If it is, return 1; otherwise, return 0.
    
    In your response, please return in the following format:
    \begin{itemize}
    \item VaxConcerns\_1: [0/1]
    \item VaxConcerns\_1.1:[0/1]
    \item \ldots
    \item VaxConcerns\_5.4:[0/1]
    \end{itemize}
    Paragraph: \ldots
\end{itemize}

The prompt for the individual prompting approach is very similar, except that at the end, it says:
\begin{itemize}[leftmargin=*]\itshape
    \item[] In your response, please only return for the VaxConcerns\_\{\} label. We will ask you about the other labels later.
\end{itemize}

\section{Resources}
\begin{itemize}[leftmargin=*]
    \item ConcernScope website: \url{https://vax.cs.duke.edu/}
    \item Source code: \url{https://gitlab.cs.duke.edu/cl619/vaccine-concern-classifier}
\end{itemize}

\section{Platform Screenshots}
\begin{figure}[H]
    \centering
    \includegraphics[width=1\linewidth, trim=0 400 0 150, clip]{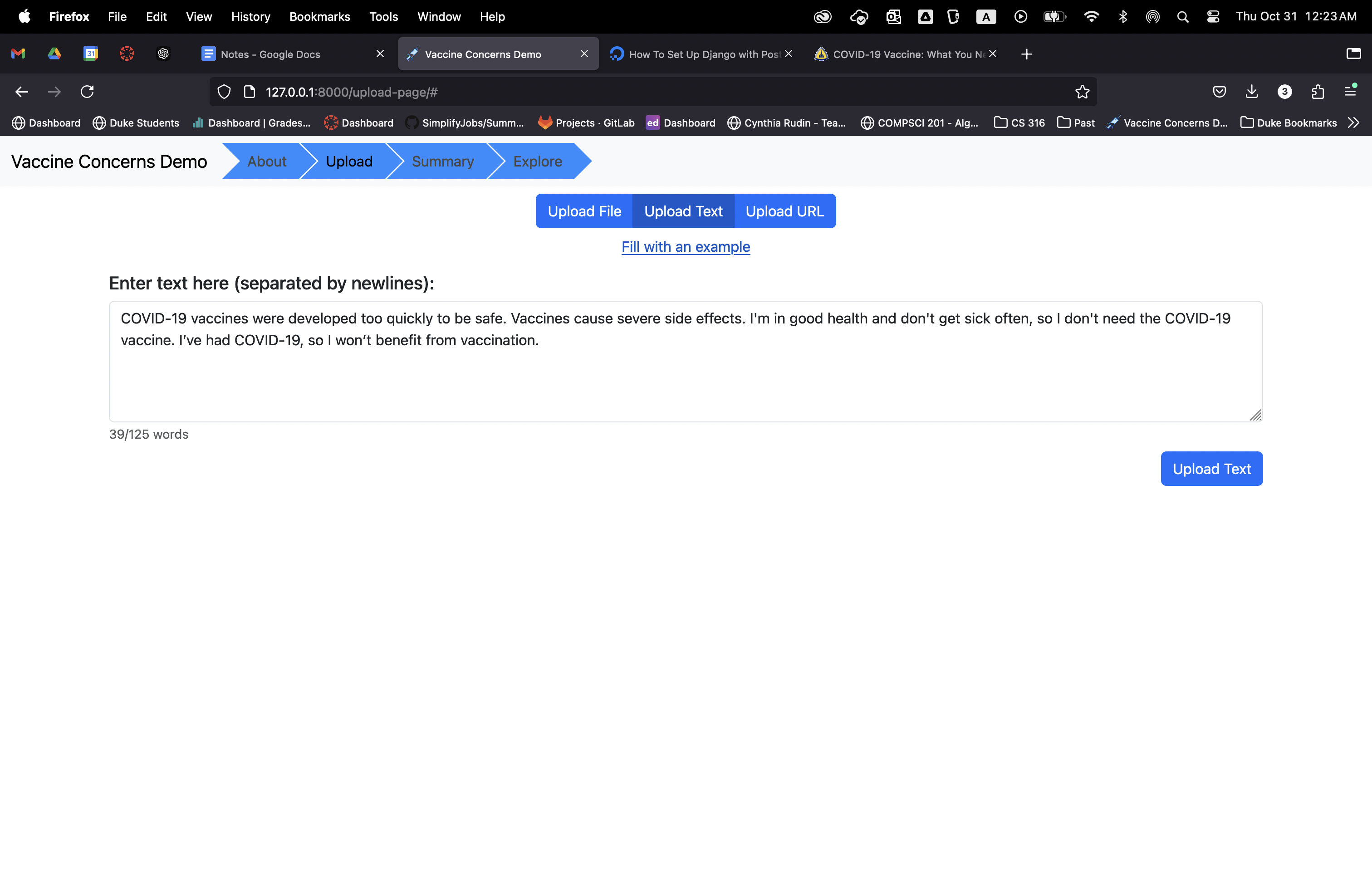}
    \caption{The Upload page of ConcernScope allows users to submit vaccine-related text for automated concern analysis. By selecting the \textit{Upload Text} tab, users can input content directly by typing or pasting text. Once submitted, the system identifies and categorizes vaccine-related concerns using the VaxConcerns taxonomy.}
    \Description{The Upload page of ConcernScope on the Upload Text tab.}
\end{figure}

\begin{figure}[H]
    \centering
    \includegraphics[width=1\linewidth, trim=0 500 0 150, clip]{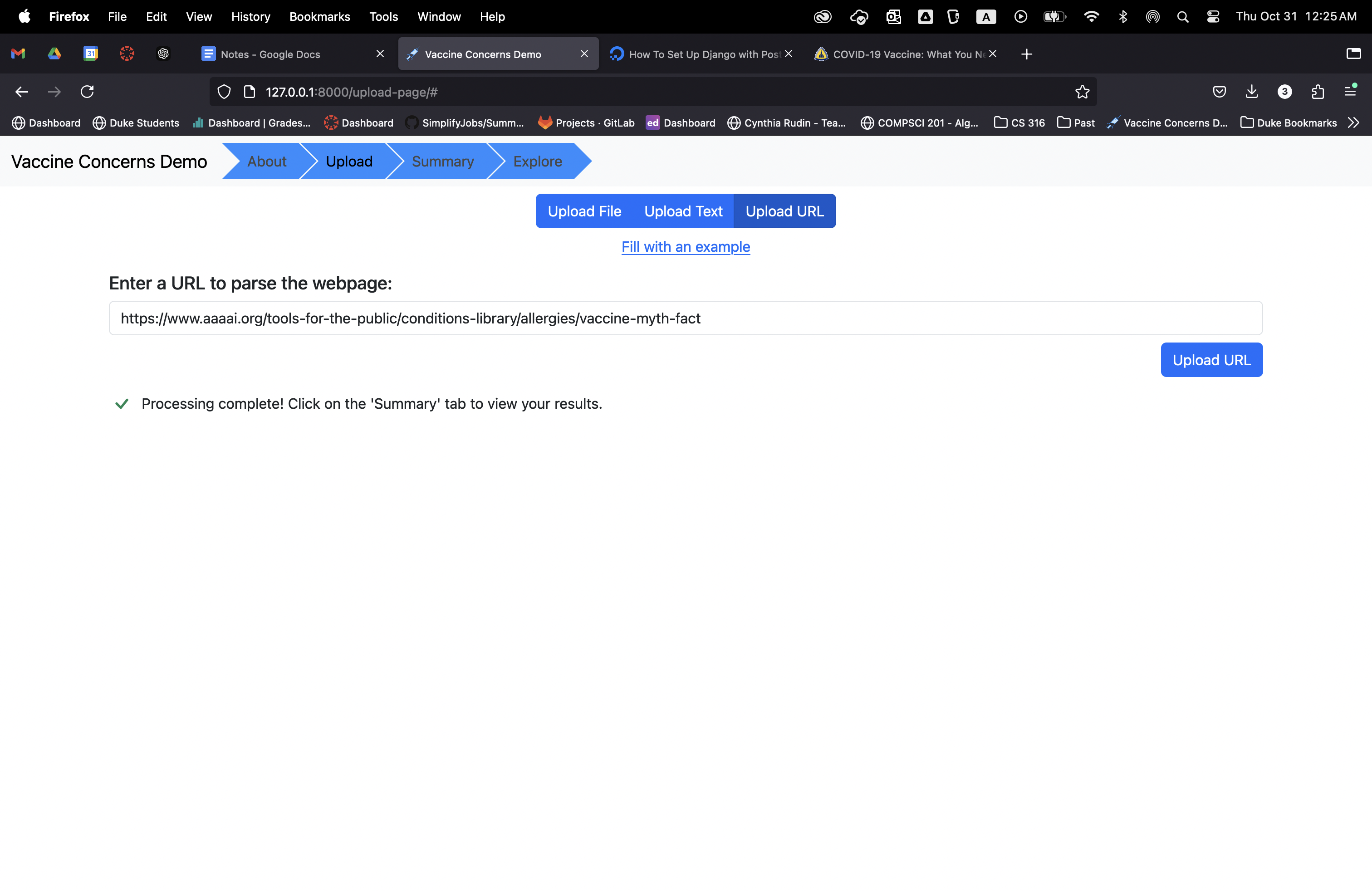}
    \caption{The Upload page of ConcernScope allows users to submit vaccine-related text for automated concern analysis. By selecting the \textit{Upload URL} tab, users can input content directly by providing a URL. Once submitted, the system identifies and categorizes vaccine-related concerns using the VaxConcerns taxonomy.}
    \Description{The Upload page of ConcernScope on the Upload URL tab.}
\end{figure}

\begin{figure}[H]
    \centering
    \includegraphics[width=0.75\linewidth, trim=100 0 100 200, clip]{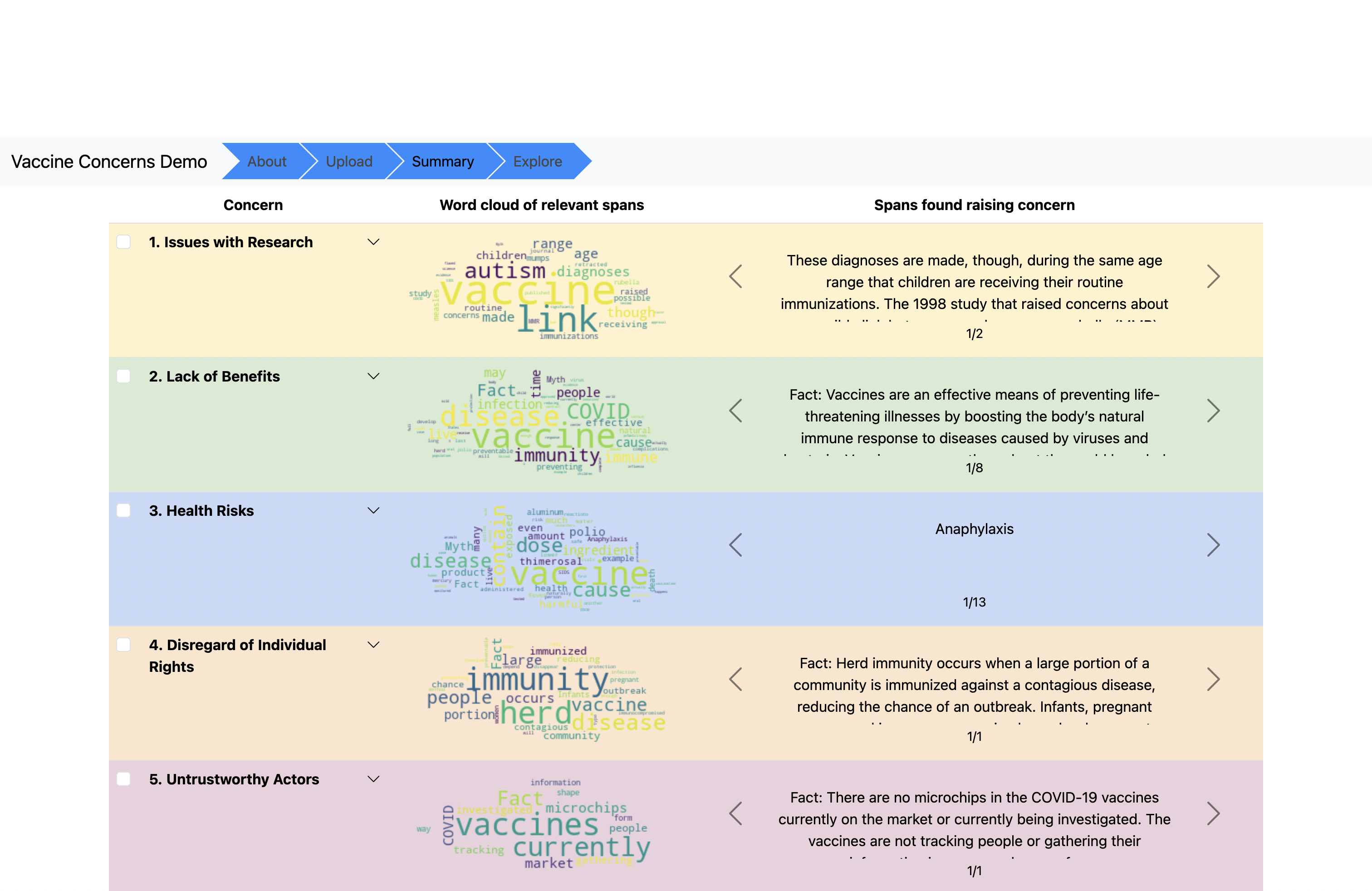}
    \caption{A full view of the Summary page of ConcernScope (as shown in Figure~\ref{fig:summary}, which allows guided analysis of common trends and key example spans. 
    Samples are organized by concern category according to the VaxConcerns taxonomy, and word clouds show keyword frequencies within each category.}
    \Description{The Summary page of ConcernScope.}
\end{figure}

\begin{figure}[H]
    \centering
    \includegraphics[width=1\linewidth, trim=0 600 0 150, clip]{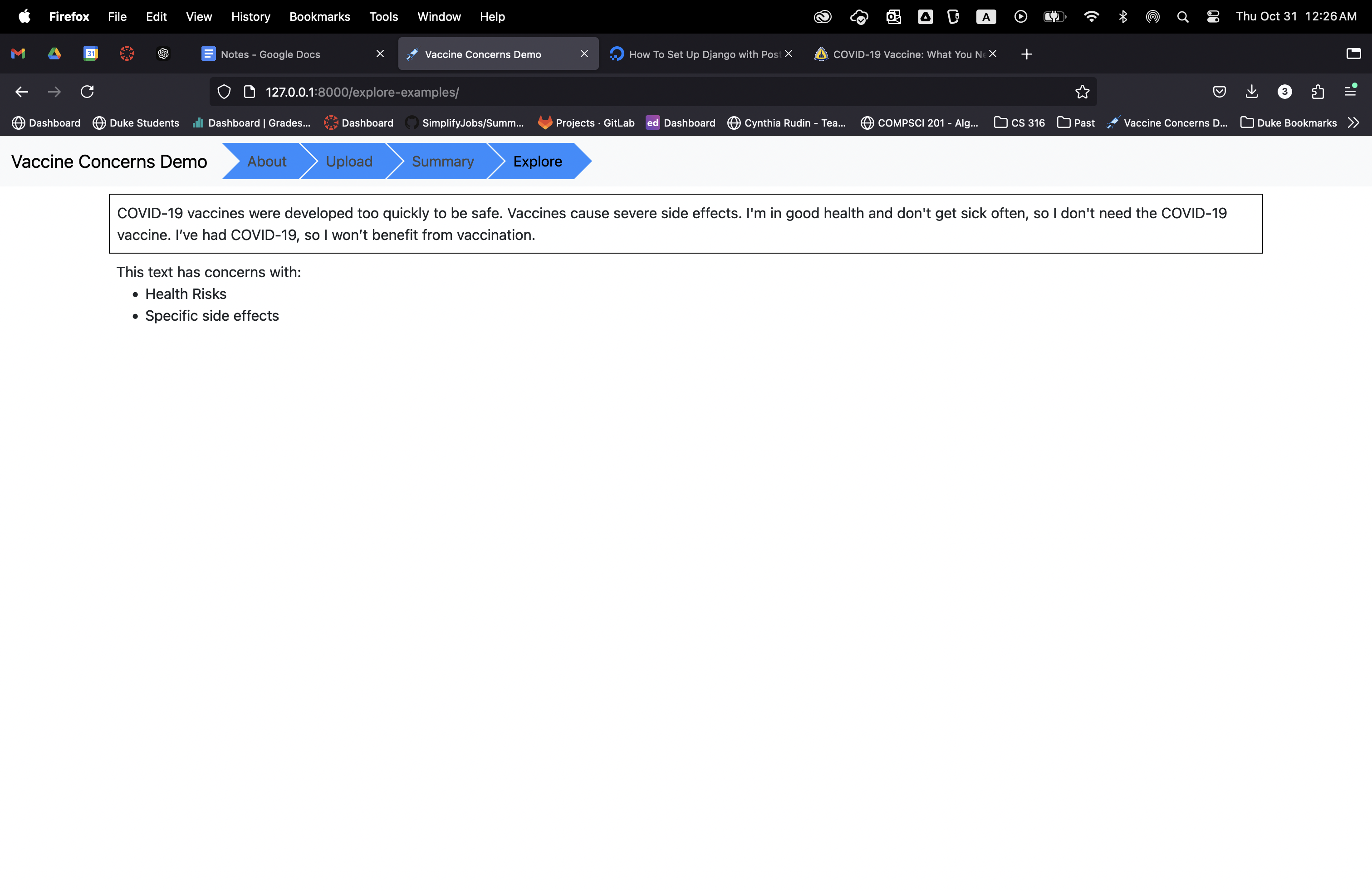}
    \caption{As shown in Figure~\ref{fig:explore}, the Explore page of ConcernScope allows users to inspect specific concerns in context. 
    When users submit text input, ConcernScope displays the vaccine-related concerns for the one sample.}
    \Description{The Explore page of ConcernScope with text input.}
\end{figure}

\begin{figure}[H]
    \centering
    \includegraphics[width=0.75\linewidth, trim=0 0 100 150, clip]{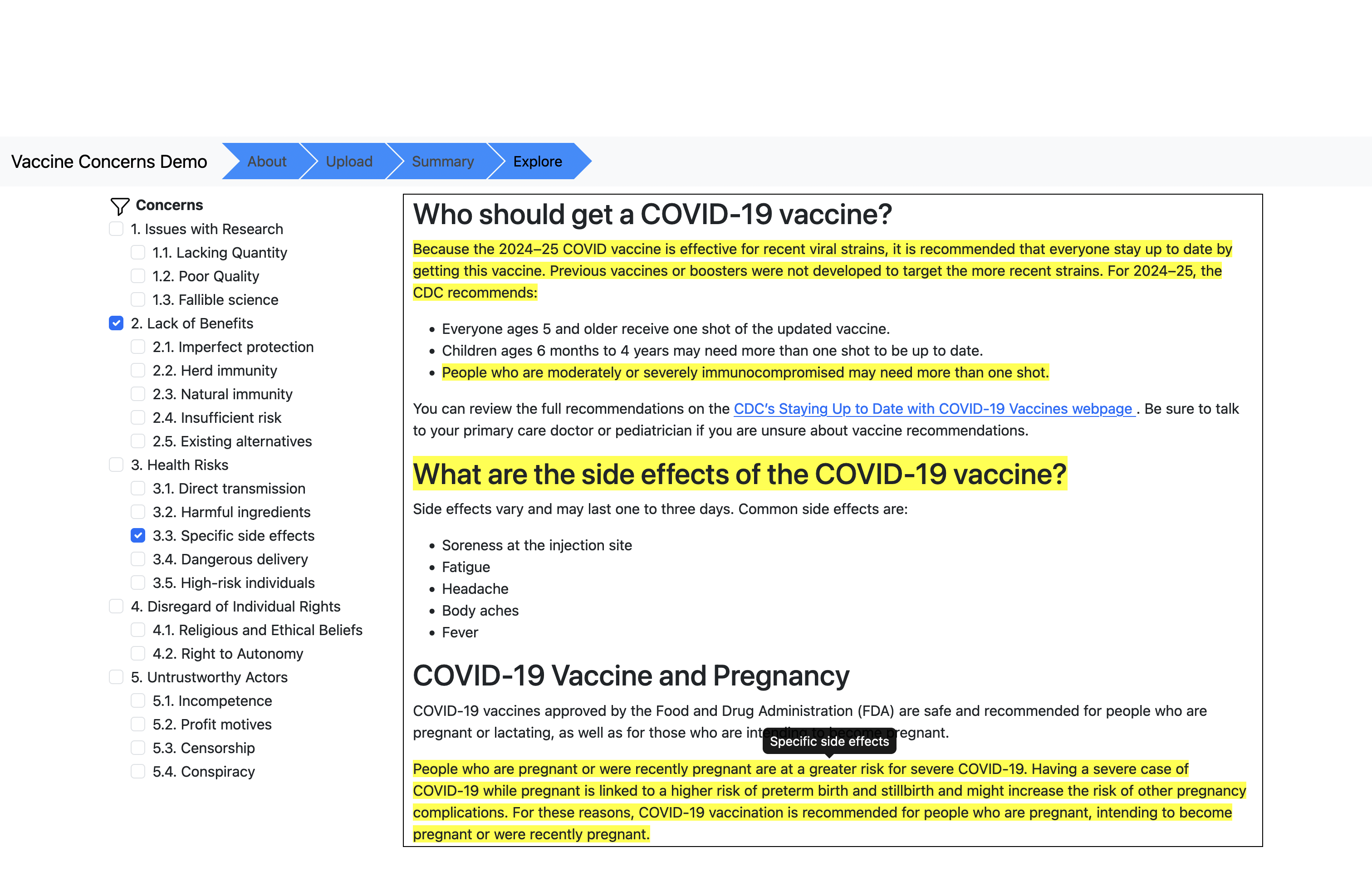}
    \caption{As shown in Figure~\ref{fig:explore}, the Explore page of ConcernScope allows users to inspect specific concerns in context. 
    Users can filter the original article text by selecting concerns from the VaxConcerns taxonomy on the left, which highlights corresponding spans within the article.}
    \Description{The Explore page of ConcernScope with website URL input.}
\end{figure}

\section{Relevance Classifier Performance}

\begin{figure}[H]
    \centering
    \includegraphics[width=0.5\linewidth]{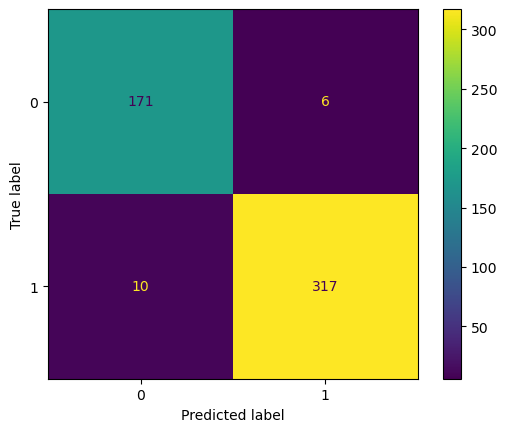}
    \caption{Confusion matrix for GPT classifications on the gold-labeled relevance dataset. 
    GPT achieved an accuracy of 0.968, precision of 0.981, recall of 0.969, and an F1 score of 0.975.}
    \Description{Evaluation of GPT on the relevance task.}
\end{figure}

\begin{figure}[H]
    \centering
    \includegraphics[width=1\linewidth]{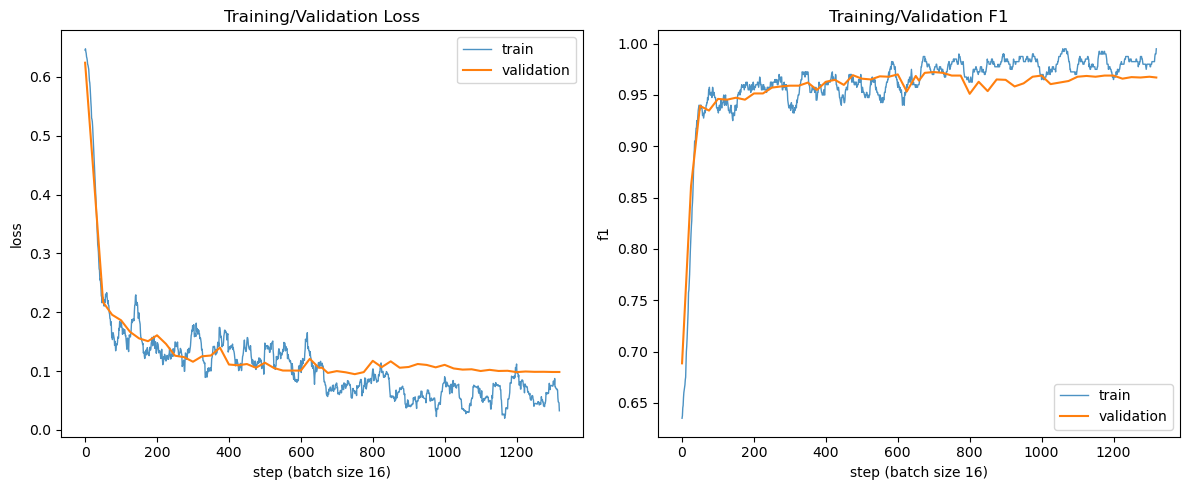}
    \caption{Training and validation loss (left) and F1 score (right) over training steps for the BERT-based relevance classifier.}
    \Description{Training and validation loss over epochs for the BERT relevance classifier.}

    \vspace{1em} 

    \includegraphics[width=0.5\linewidth]{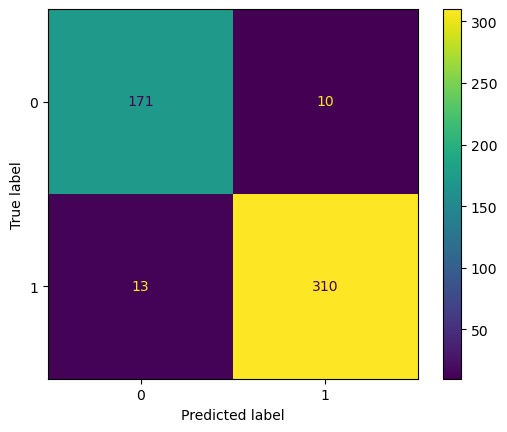}
    \caption{Confusion matrix for BERT classifications on the gold-labeled relevance dataset. 
    The model achieved an accuracy of 0.954, precision of 0.969, recall of 0.960, and an F1 score of 0.964.}
    \Description{Evaluation of the BERT classifier on the relevance task.}
\end{figure}

\newpage
\section{Multilabel Classifier Performance}

\begin{center}
\captionof{table}{Classification report for GPT-4 annotations using prompts that ask for each VaxConcerns Taxonomy label individually.}
\begin{adjustbox}{max width=\textwidth}
\begin{tabular}{lcccc}
\toprule
\textbf{Label} & \textbf{Precision} & \textbf{Recall} & \textbf{F1-score} & \textbf{Support} \\
\midrule
1       & 0.67 & 0.63 & 0.65 & 19 \\
1.1     & 1.00 & 0.25 & 0.40 & 8 \\
1.2     & 0.73 & 0.67 & 0.70 & 12 \\
1.3     & 0.00 & 0.00 & 0.00 & 0 \\
2       & 1.00 & 0.78 & 0.88 & 18 \\
2.1     & 1.00 & 0.69 & 0.82 & 13 \\
2.2     & 0.00 & 0.00 & 0.00 & 0 \\
2.3     & 1.00 & 1.00 & 1.00 & 1 \\
2.4     & 1.00 & 0.50 & 0.67 & 2 \\
2.5     & 1.00 & 1.00 & 1.00 & 2 \\
3       & 0.92 & 0.72 & 0.81 & 47 \\
3.1     & 0.00 & 0.00 & 0.00 & 0 \\
3.2     & 1.00 & 1.00 & 1.00 & 11 \\
3.3     & 0.77 & 0.29 & 0.43 & 34 \\
3.4     & 0.25 & 0.33 & 0.29 & 3 \\
3.5     & 0.12 & 0.50 & 0.20 & 2 \\
4       & 0.88 & 0.70 & 0.78 & 10 \\
4.1     & 1.00 & 1.00 & 1.00 & 2 \\
4.2     & 0.67 & 0.80 & 0.73 & 5 \\
5       & 0.93 & 0.82 & 0.87 & 33 \\
5.1     & 1.00 & 1.00 & 1.00 & 1 \\
5.2     & 0.79 & 0.88 & 0.83 & 17 \\
5.3     & 0.67 & 0.40 & 0.50 & 5 \\
5.4     & 0.50 & 0.50 & 0.50 & 4 \\
\midrule
micro avg    & 0.47 & 0.70 & 0.56 & 249 \\
macro avg    & 0.42 & 0.67 & 0.46 & 249 \\
weighted avg & 0.65 & 0.70 & 0.62 & 249 \\
samples avg  & 0.46 & 0.68 & 0.51 & 249 \\
\bottomrule
\end{tabular}
\end{adjustbox}
\end{center}

\begin{table}[H]
\centering
\caption{Classification report for GPT-4 annotations using prompts that ask for each VaxConcerns Taxonomy label individually.}
\begin{adjustbox}{max width=1\textwidth}
\begin{tabular}{lcccc}
\toprule
\textbf{Label} & \textbf{Precision} & \textbf{Recall} & \textbf{F1-score} & \textbf{Support} \\
\midrule
1       & 0.34 & 0.89 & 0.49 & 19 \\
1.1     & 0.67 & 0.50 & 0.57 & 8 \\
1.2     & 0.36 & 0.75 & 0.49 & 12 \\
1.3     & 0.00 & 0.00 & 0.00 & 0 \\
2       & 0.71 & 0.28 & 0.40 & 18 \\
2.1     & 0.86 & 0.46 & 0.60 & 13 \\
2.2     & 0.00 & 0.00 & 0.00 & 0 \\
2.3     & 0.33 & 1.00 & 0.50 & 1 \\
2.4     & 0.33 & 1.00 & 0.50 & 2 \\
2.5     & 0.67 & 1.00 & 0.80 & 2 \\
3       & 0.86 & 0.79 & 0.82 & 47 \\
3.1     & 0.00 & 0.00 & 0.00 & 0 \\
3.2     & 0.62 & 0.91 & 0.74 & 11 \\
3.3     & 0.81 & 0.50 & 0.62 & 34 \\
3.4     & 0.20 & 0.67 & 0.31 & 3 \\
3.5     & 0.05 & 0.50 & 0.10 & 2 \\
4       & 0.37 & 0.70 & 0.48 & 10 \\
4.1     & 0.67 & 1.00 & 0.80 & 2 \\
4.2     & 0.50 & 0.80 & 0.62 & 5 \\
5       & 0.71 & 0.76 & 0.74 & 33 \\
5.1     & 0.06 & 1.00 & 0.12 & 1 \\
5.2     & 0.60 & 0.88 & 0.71 & 17 \\
5.3     & 0.20 & 0.80 & 0.32 & 5 \\
5.4     & 0.16 & 1.00 & 0.28 & 4 \\
\midrule
micro avg    & 0.47 & 0.70 & 0.56 & 249 \\
macro avg    & 0.42 & 0.67 & 0.46 & 249 \\
weighted avg & 0.65 & 0.70 & 0.62 & 249 \\
samples avg  & 0.46 & 0.68 & 0.51 & 249 \\
\bottomrule
\end{tabular}
\end{adjustbox}
\end{table}

\begin{figure}[H]
    \centering
    \includegraphics[width=0.75\linewidth]{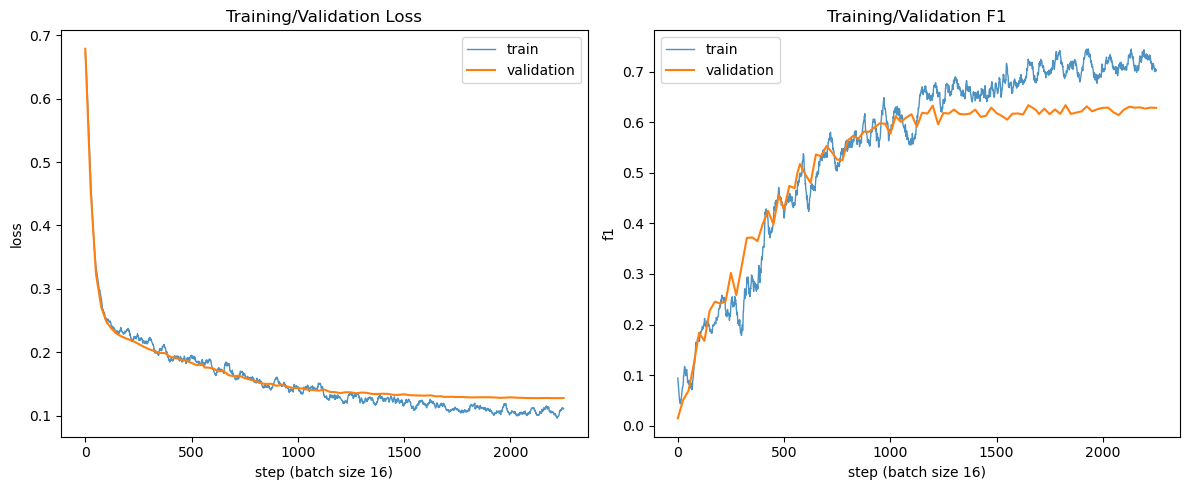}
    \caption{Training/validation loss and F1 curves for the multilabel GPT classifier, which classifies text according to the VaxConcerns Taxonomy.}
\end{figure}

\begin{table}[H]
\centering
\caption{Classification report for the baseline BERT-based multilabel classification model. It weights each class equally and classifies according to the VaxConcerns Taxonomy.}
\begin{adjustbox}{max width=1\textwidth}
\begin{tabular}{lcccc}
\toprule
\textbf{Label} & \textbf{Precision} & \textbf{Recall} & \textbf{F1-score} & \textbf{Support} \\
\midrule
0   & 0.70 & 0.37 & 0.48 & 19 \\
1   & 0.00 & 0.00 & 0.00 & 8 \\
2   & 0.80 & 0.33 & 0.47 & 12 \\
3   & 0.00 & 0.00 & 0.00 & 0 \\
4   & 1.00 & 0.56 & 0.71 & 18 \\
5   & 1.00 & 0.15 & 0.27 & 13 \\
6   & 0.00 & 0.00 & 0.00 & 0 \\
7   & 0.00 & 0.00 & 0.00 & 1 \\
8   & 1.00 & 0.50 & 0.67 & 2 \\
9   & 0.00 & 0.00 & 0.00 & 2 \\
10  & 0.93 & 0.83 & 0.88 & 47 \\
11  & 0.00 & 0.00 & 0.00 & 0 \\
12  & 0.89 & 0.73 & 0.80 & 11 \\
13  & 1.00 & 0.09 & 0.16 & 34 \\
14  & 0.00 & 0.00 & 0.00 & 3 \\
15  & 0.00 & 0.00 & 0.00 & 2 \\
16  & 0.86 & 0.60 & 0.71 & 10 \\
17  & 0.00 & 0.00 & 0.00 & 2 \\
18  & 0.57 & 0.80 & 0.67 & 5 \\
19  & 0.92 & 0.73 & 0.81 & 33 \\
20  & 0.00 & 0.00 & 0.00 & 1 \\
21  & 0.86 & 0.71 & 0.77 & 17 \\
22  & 0.00 & 0.00 & 0.00 & 5 \\
23  & 0.00 & 0.00 & 0.00 & 4 \\
\midrule
micro avg    & 0.88 & 0.48 & 0.62 & 249 \\
macro avg    & 0.44 & 0.27 & 0.31 & 249 \\
weighted avg & 0.80 & 0.48 & 0.56 & 249 \\
samples avg  & 0.74 & 0.49 & 0.56 & 249 \\
\bottomrule
\end{tabular}
\end{adjustbox}
\end{table}

\begin{figure}[H]
    \centering
    \includegraphics[width=0.7\linewidth]{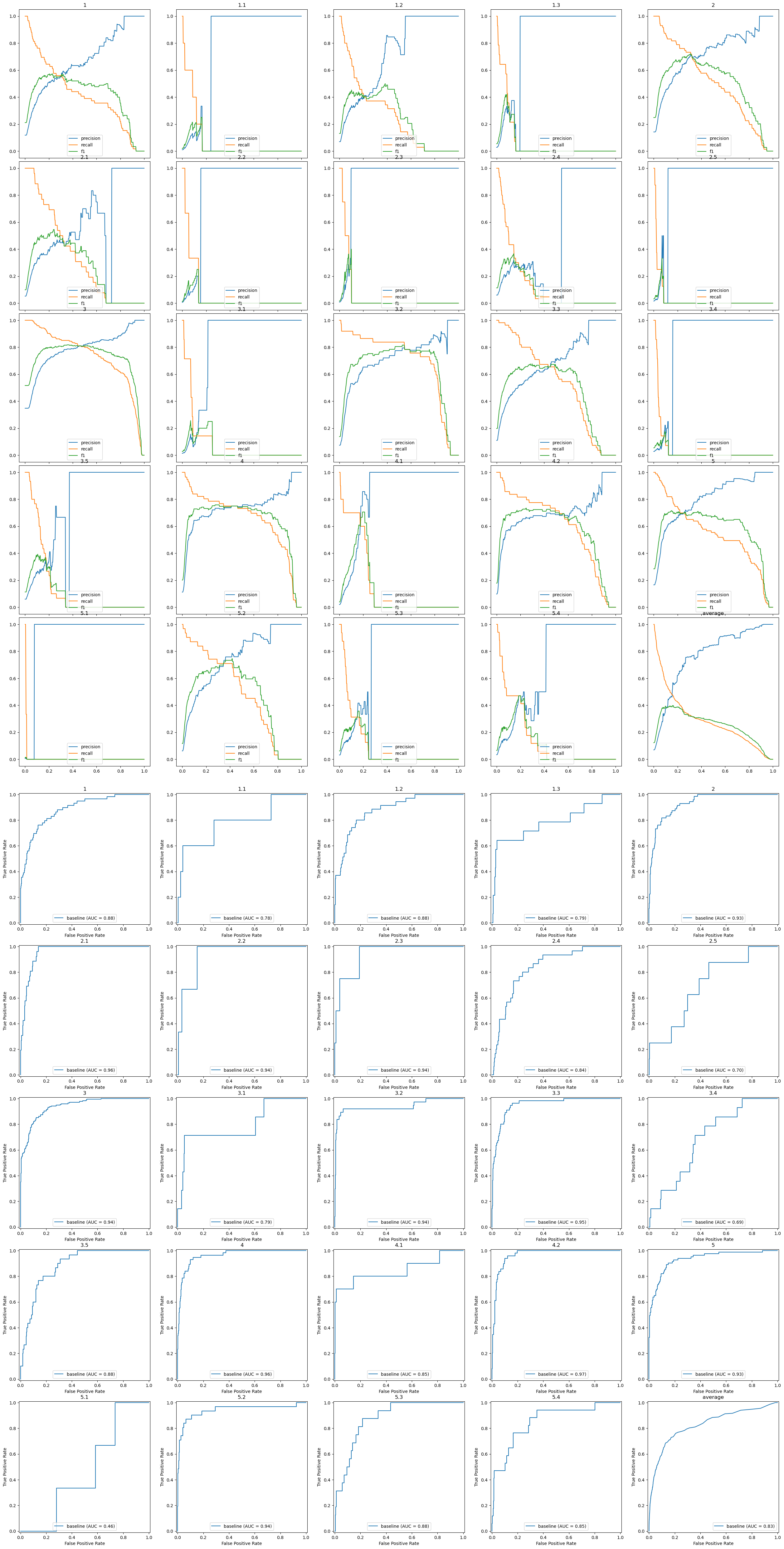}
    \caption{Evaluation of the baseline multilabel classifier trained with equal class weights. The top half of the figure contains per-label precision, recall, and F1 metrics, and the bottom half has per-label ROC-AUC curves.}
    \Description{Evaluation of the baseline multilabel classifier.}
\end{figure}

\begin{table}[H]
\centering
\caption{Classification report for the weighted BERT-based multilabel classification model. It weights each class using the log1p scheme and classifies according to the VaxConcerns Taxonomy.}
\begin{adjustbox}{max width=\textwidth}
\begin{tabular}{lcccc}
\toprule
\textbf{Label} & \textbf{Precision} & \textbf{Recall} & \textbf{F1-score} & \textbf{Support} \\
\midrule
0   & 0.73 & 0.42 & 0.53 & 19 \\
1   & 0.00 & 0.00 & 0.00 & 8 \\
2   & 0.88 & 0.58 & 0.70 & 12 \\
3   & 0.00 & 0.00 & 0.00 & 0 \\
4   & 1.00 & 0.50 & 0.67 & 18 \\
5   & 0.75 & 0.23 & 0.35 & 13 \\
6   & 0.00 & 0.00 & 0.00 & 0 \\
7   & 1.00 & 1.00 & 1.00 & 1 \\
8   & 1.00 & 0.50 & 0.67 & 2 \\
9   & 0.50 & 0.50 & 0.50 & 2 \\
10  & 0.89 & 0.72 & 0.80 & 47 \\
11  & 0.00 & 0.00 & 0.00 & 0 \\
12  & 0.82 & 0.82 & 0.82 & 11 \\
13  & 0.92 & 0.32 & 0.48 & 34 \\
14  & 0.00 & 0.00 & 0.00 & 3 \\
15  & 0.00 & 0.00 & 0.00 & 2 \\
16  & 0.82 & 0.90 & 0.86 & 10 \\
17  & 1.00 & 0.50 & 0.67 & 2 \\
18  & 0.80 & 0.80 & 0.80 & 5 \\
19  & 0.88 & 0.67 & 0.76 & 33 \\
20  & 0.00 & 0.00 & 0.00 & 1 \\
21  & 0.81 & 0.76 & 0.79 & 17 \\
22  & 0.00 & 0.00 & 0.00 & 5 \\
23  & 0.00 & 0.00 & 0.00 & 4 \\
\midrule
micro avg    & 0.84 & 0.53 & 0.65 & 249 \\
macro avg    & 0.53 & 0.38 & 0.43 & 249 \\
weighted avg & 0.78 & 0.53 & 0.62 & 249 \\
samples avg  & 0.70 & 0.56 & 0.60 & 249 \\
\bottomrule
\end{tabular}
\end{adjustbox}
\end{table}

\begin{figure}[H]
    \centering
    \includegraphics[width=0.64\linewidth]{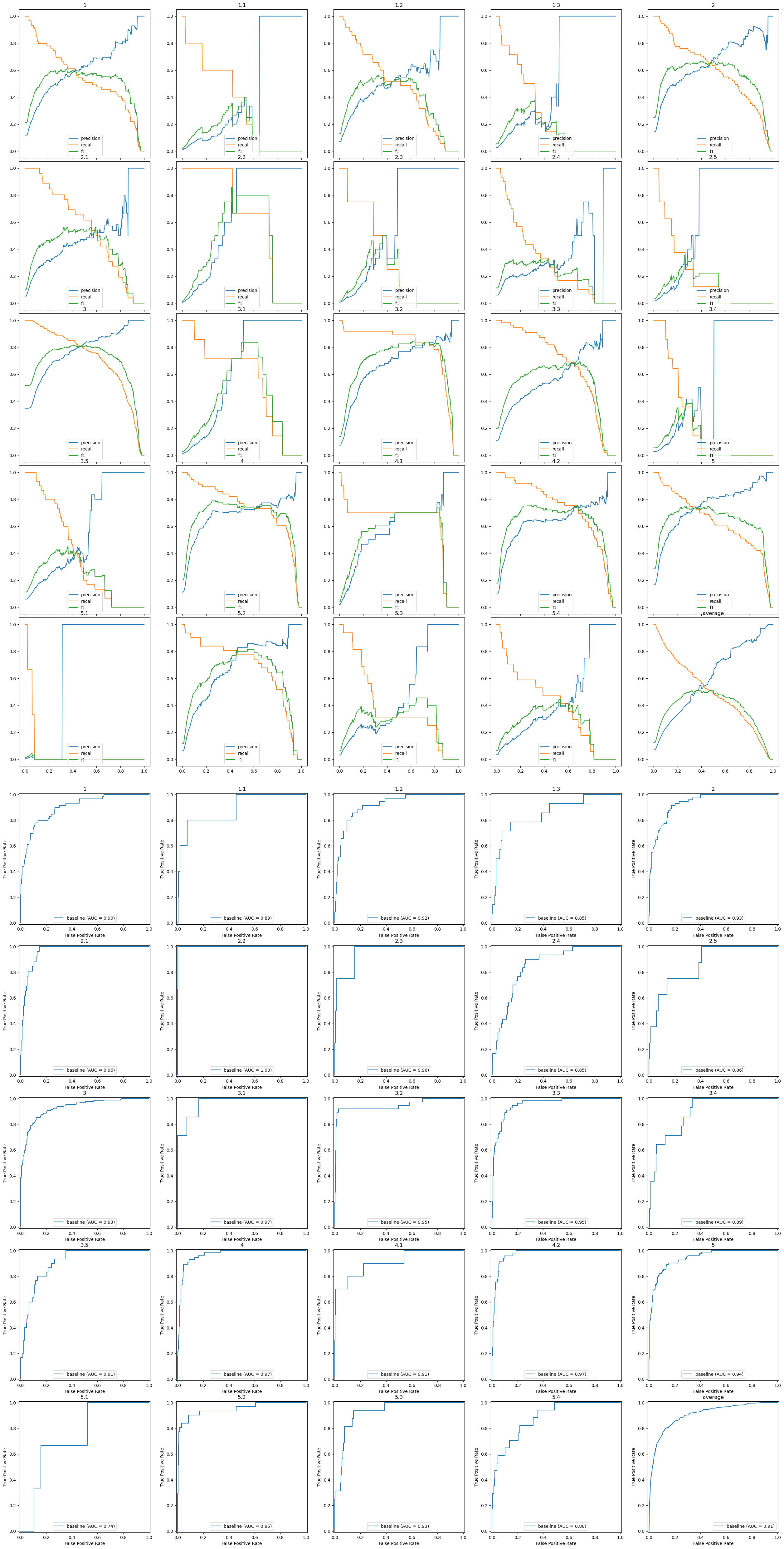}
    \caption{Evaluation of the weighted multilabel classifier trained with class weighting using the log1p scheme. The top half of the figure contains per-label precision, recall, and F1 metrics, and the bottom half has per-label ROC-AUC curves.}
    \Description{Evaluation of the weighted multilabel classifier.}
\end{figure}

\end{document}